\documentclass[times,twocolumn,final]{elsarticle}
\usepackage{cag}
\usepackage{framed,multirow}

\usepackage{amssymb}
\usepackage{latexsym}

\usepackage{url}
\usepackage{xcolor}
\definecolor{newcolor}{rgb}{.8,.349,.1}

\usepackage{hyperref}

\usepackage[switch,pagewise]{lineno} %Required by command \linenumbers below

\usepackage{subfigure}
\usepackage{caption}

\newcommand{\zjq}[1]{{\color{black}#1}}

\usepackage{xspace}
\usepackage{enumitem}

\usepackage[capitalize]{cleveref}
\crefname{section}{Sec.}{Secs.}
\Crefname{section}{Section}{Sections}
\Crefname{table}{Table}{Tables}
\crefname{table}{Tab.}{Tabs.}

\journal{Computers \& Graphics}

\begin{document}

\verso{Preprint Submitted for review}

\begin{frontmatter}

\title{HoughLaneNet: Lane Detection with Deep Hough Transform and Dynamic Convolution}%
% \tnotetext[tnote1]{Only capitalize first
% word and proper nouns in the title.}

% -----------------------------------------------------------
\author[1]{Jia-Qi~Zhang\fnref{fn1}}
    
\author[1]{Hao-Bin~Duan\fnref{fn1}}

\author[1]{Jun-Long~Chen\fnref{fn1}}
\fntext[fn1]{E-mails:\{zhangjiaqi79, duanhb, junlong2000\}@buaa.edu.cn.}  

\author[2]{Ariel Shamir\fnref{fn2}}
\fntext[fn2]{E-mail:arik@idc.ac.il.}  

\author[1,3]{Miao Wang\corref{cor1}}
% \cortext[cor1]{Corresponding author: 
%   Tel.: +86-189-0000-0000;  
%   fax: +0-000-000-0000;}
\cortext[cor1]{Corresponding author.}
\emailauthor{miaow@buaa.edu.cn}{Miao Wang}
%\ead{example@email.com}

\address[1]{The State Key Laboratory of Virtual Reality Technology and Systems, Beihang University, Beijing, 100191, China}
\address[2]{The Efi Arazi School of Computer Science, Reichman University, Herzliya, 4610000, Israel}
\address[3]{The Zhongguancun Laboratory, Beijing, 100195, China}

% -----------------------------------------------------------

%\received{1 February 2017}
\received{\today}
%%%% Do not use the below for submitted manuscripts
%\finalform{28 March 2017}
\accepted{19 August 2023}
%\availableonline{15 May 2017}
%\communicated{S. Sarkar}

\begin{abstract}
%%%
The task of lane detection has garnered considerable attention in the field of autonomous driving due to its complexity. Lanes can present difficulties for detection, as they can be narrow, fragmented, and often obscured by heavy traffic. However, it has been observed that the lanes have a geometrical structure that resembles a straight line, leading to improved lane detection results when utilizing this characteristic. To address this challenge, we propose a hierarchical Deep Hough Transform (DHT) approach that combines all lane features in an image into the Hough parameter space. Additionally, we refine the point selection method and incorporate a Dynamic Convolution Module to effectively differentiate between lanes in the original image. Our network architecture comprises a backbone network, either a ResNet or Pyramid Vision Transformer, a Feature Pyramid Network as the neck to extract multi-scale features, and a hierarchical DHT-based feature aggregation head to accurately segment each lane. By utilizing the lane features in the Hough parameter space, the network learns dynamic convolution kernel parameters corresponding to each lane, allowing the Dynamic Convolution Module to effectively differentiate between lane features. Subsequently, the lane features are fed into the feature decoder, which predicts the final position of the lane. Our proposed network structure demonstrates improved performance in detecting heavily occluded or worn lane images, as evidenced by our extensive experimental results, which show that our method outperforms or is on par with state-of-the-art techniques.

%%%%
\end{abstract}


\end{frontmatter}

% \linenumbers


\section{Introduction}

An advanced driver-assistance system (ADAS) leverages various automation technologies to enable different degrees of autonomous driving. These techniques include the detection of various objects such as pedestrians~\cite{AngelovaKVOF15, Pedestrian_MaoXJC17, Parallelized_ZhouZ16}, traffic signs~\cite{TrafficSignsZhuLMH17, TrafficDetectionZhuLZHLH16, panoramas_SongFHZT19}, and lanes~\cite{endtoendinstanceseg, garnett20193dlanenet, SpinNet_FanWHLM19} through the use of sensors and cameras. Lane detection plays a crucial role in ensuring the safety of autonomous driving, which provides guidance for adjusting driving direction, maintaining cruise stability, and most importantly, ensures safety by minimizing the risk of collisions through lane preservation assistance, departure warnings, and centerline-based lateral control technologies~\cite{linecnn, lu2020super, centerline}. Lane detection techniques are also beneficial in creating high-definition maps of highways~\cite{homayounfar2020dagmapper}.

\begin{figure}[t!]
   \centering
   \includegraphics[width=0.99\linewidth]{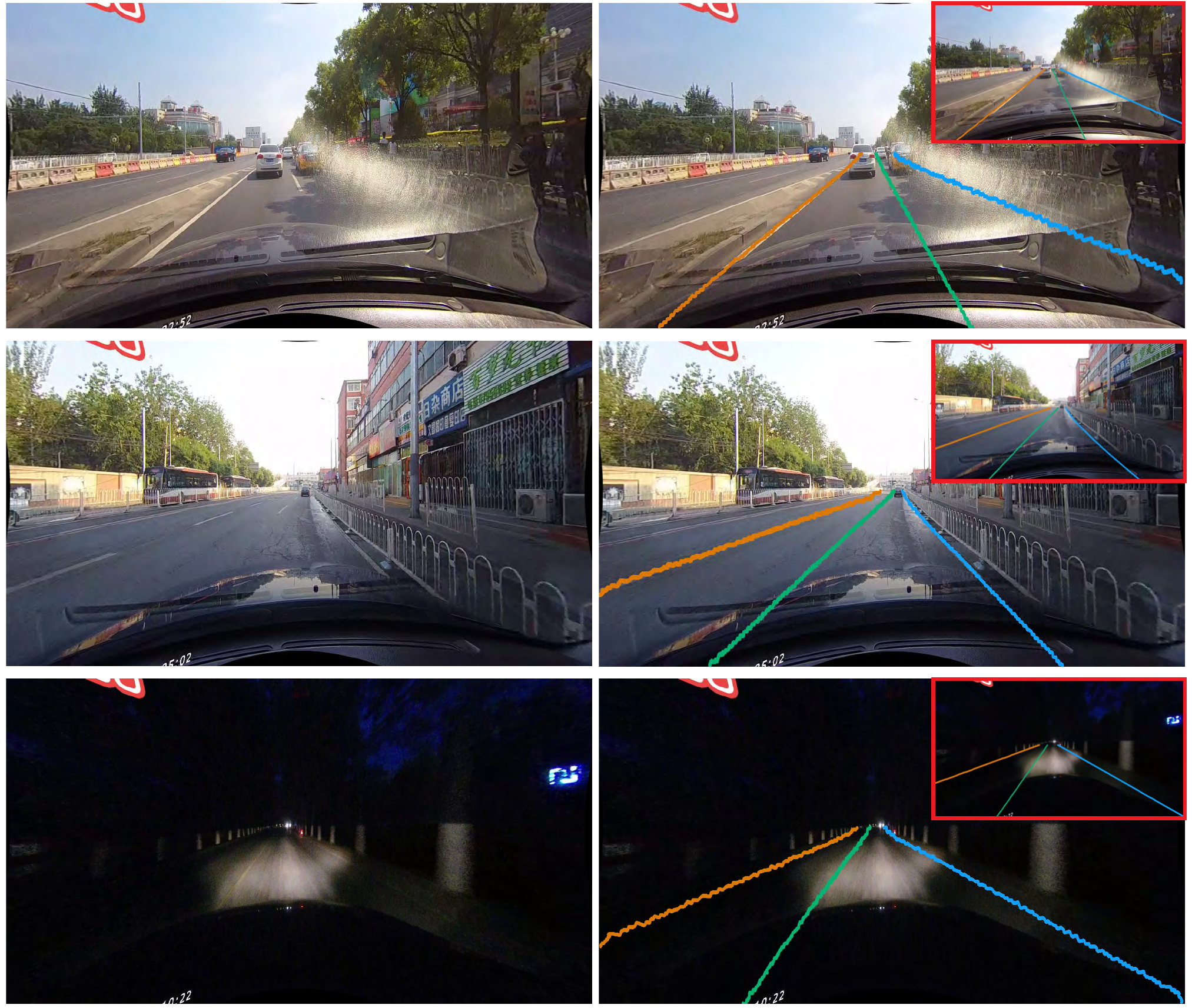}
   \caption{Lane detection results for complex scenes. We show examples of detection results of dazzle, worn and dark lanes, and use different colors to represent different lanes. The left column shows the input images for lane detection, the right column shows the predicted lanes using our method, where the top-right corner displays the thumbnails of ground truth results.}
   \label{fig:example_hard}
\end{figure}
% 现有方法，挑战，问题
% obscure, occluded, or discontinuous

Current monocular lane detection methods detect lanes on roads with a camera installed on the vehicle. The state-of-the-art methods~\cite{LaneATT,condlanenet,zheng2022clrnet} are able to achieve high accuracy in simple scenarios containing mostly straight lines which are rarely occluded, under well-lit conditions, such as the ones in the TuSimple dataset~\cite{TuSimple} that consists of highway road images taken during daytime. However, such testing examples do not generalize well to real-world scenarios in streets which involve other visually-complicated objects such as buildings, pedestrians, and sidewalks. In real streets, challenging cases including worn lanes, discontinuous lanes, and lanes under extreme lighting conditions can be quite common.

In order to tackle complex scenarios, a robust lane detection model must have the capability to detect complete lanes even in the presence of weak, limited, or absent visual features, as demonstrated in Figure~\ref{fig:example_hard}. This requires the model to leverage global visual features and have prior knowledge of the potential shape of the lane lines. To address these challenges, researchers have proposed various techniques to gather global visual information. For example, Lee et al.~\cite{lee2022robust} proposed an Expanded Self Attention (ESA) module to overcome occlusion problems by using global contextual information to calculate the confidence of occluded lanes. Tabelini et al.~\cite{LaneATT} presented an anchor-based network that regresses the coordinates of each row of points by projecting the lane onto a straight line. Zheng et al.~\cite{zheng2021resa} developed a method that enables each pixel to collect global information through a recurrent shifting of sliced feature maps in both vertical and horizontal directions, thus improving lane detection performance in complex scenes with weak appearance cues. Our proposed method combines the use of line-based geometry prior with Deep Hough Transform and dynamic convolution module to aggregate global features and select lane instances, resulting in a highly effective and promising solution for lane detection.

% 这一段在较高的层面介绍我们方法的总体概览

We present HoughLaneNet, a deep neural network designed to exploit the commonality that lanes are primarily straight lines, allowing for the better gathering of global information in the image and performing instance proposal, which is achieved through the use of Hough Transform, a well-established technique for mapping line features in terms of potential line positions and orientations. The resulting Hough feature space effectively combines scattered local features across the image and provides instance-specific convolution for subsequent pixel-level lane segmentation. Despite relying on a prior of straight lane shape, our method can still handle curved lanes. This is because Hough transform is mainly used to enhance feature clustering and lane instance selection, while the final lane extraction is still performed by pixel segmentation. Furthermore, our method is capable of detecting varying amounts of lanes within an image by adjusting the detection threshold.

% 这一段更详细介绍一下Line based feature aggreation 是怎么完成的

Our network design leverages the geometry prior to gathering hierarchical line-shaped information and improves the global feature fusion of lane detection, especially for lanes with weak and subtle visual cues. The network first extracts deep features from lane images, and then aggregates these features through a hierarchical Hough Transform at three scales from coarse to fine. The Hough Transform maps each feature into a two-dimensional parameter space ($R, \Theta$), as proposed for line detection task in Deep Hough Transform (DHT)~\cite{lin2020deep, 21PAMI-dht}. The transformed features from various scales are then rescaled and concatenated, resulting in a Hough map that highlights the occurrence of the lanes in the original image, enabling the network to accurately identify the number and approximate location of each lane instance within the image.

% 这一段介绍动态卷积与Hough features的关系

The \zjq{Hough} map can detect the number and rough location of the lanes in the image, but it cannot provide precise lane positions. To address this issue, we design a dynamic convolution module that can adaptively segment the lane features in the Euclidean space, using the \zjq{Hough} feature as a prior knowledge. After instance lane segmentation, we combine the lane prediction module to output the lane positions. In addition, the network selects deep features in the parameter space based on the positions of the Hough map points. We evaluate two \zjq{Hough} point selection methods and demonstrate their ability to detect varying numbers of lanes based on the set threshold. Figure~\ref{fig:example_hard} shows the prediction results of HoughLaneNet for different complex scenes.

The main contributions of this paper include:

\begin{itemize}

\item We introduce a hierarchical Deep Hough Transform approach for lane detection that leverages the inductive bias of the commonly straight lane shape to perform effective feature aggregation and instance proposal.

\item We introduce an optimized lane instance selection method and dynamic convolution module for improved lane segmentation, complemented with a reverse Hough transform loss to reduce residual error in ground-truth Hough maps.

\item We perform extensive experimentation and demonstrated that our model achieved state-of-the-art lane detection performance on three benchmark datasets.

\end{itemize}

We evaluate our model on three popular benchmarks: TuSimple~\cite{TuSimple}, CULane~\cite{SCNN}, and LLAMAS~\cite{llamas2019}. The proposed HoughLaneNet achieves competitive results compared with current state-of-the-art methods on all datasets, obtaining the highest accuracy of up to $96.93\%$ on TuSimple, a high F1 value of $95.49$ on LLAMAS, and an F1 value of $79.81$ on CULane. The code will become publicly available after the paper’s acceptance.

\section{Related Work}\label{sec:relatedwork}

Lane detection~\cite{liang2020lane} mainly relied on classical computer vision. Narote et al.~\cite{NAROTE2018216} provided a comprehensive review of traditional vision-based lane detection techniques, including Hough transform approaches~\cite{hough1, zheng2018improved, mao2012lane}, Inverse Perspective Mapping (IPM)~\cite{ipm1, ipm2}, Sobel~\cite{sobeledge1, sobeledge2} and canny edges~\cite{cannyedge1, cannyedge2}. However, these  methods could degenerate in complex environments. Ever since the introduction of deep learning to computer vision, various learning-based approaches emerged, which have achieved significantly better results for the lane detection task.

\textbf{Classification-based Methods.\thinspace} 
Many recent techniques adopt row-wise classification to predict the horizontal coordinates of lanes. For each row, only one cell with the highest probability is determined as a lane point, and the whole lane is detected by repeating this process. 
% and repeats this process for each lane. 
A post-processing step is required to further reconstruct all discrete points into structured lanes, which is usually completed using shape parameter regression. Qin et al.~\cite{ultrafast} proposed a row-based selection method to extract global lane features that achieves remarkable speed and precision. Yoo et al.~\cite{endtoend2020} proposed an end-to-end lane marker vertex prediction method based on a novel  horizontal reduction module by utilizing the original shape of lane markers without the need for sophisticated post-processing. Both of the above methods achieved good results on the TuSimple and CULane datasets. Liu et al.~\cite{condlanenet} introduced a heatmap to distinguish different lane instances, and applied segmentation for each lane with dynamic convolution. These row-wise classification approaches are simple to implement and 
perform fast. However, they do not use the prior knowledge of the slender line structure of lanes, thus easily fail in scenes with serious occlusion and discontinuity. Zheng et al.~\cite{zheng2022clrnet} proposed a cross-layer refinement method that uses lane priors to segment lanes in a coarse-to-fine manner, and extracts global contextual information from lane priors to further improve the results with a novel loss that minimizes localization error. Although our model is also based on pixel-level classification, we inherently exploit geometry prior of lanes with our proposed feature aggregation and instance proposal modules based on Deep Hough Transform.

\begin{figure}[t!]
   \centering
   \includegraphics[width=0.98\linewidth]{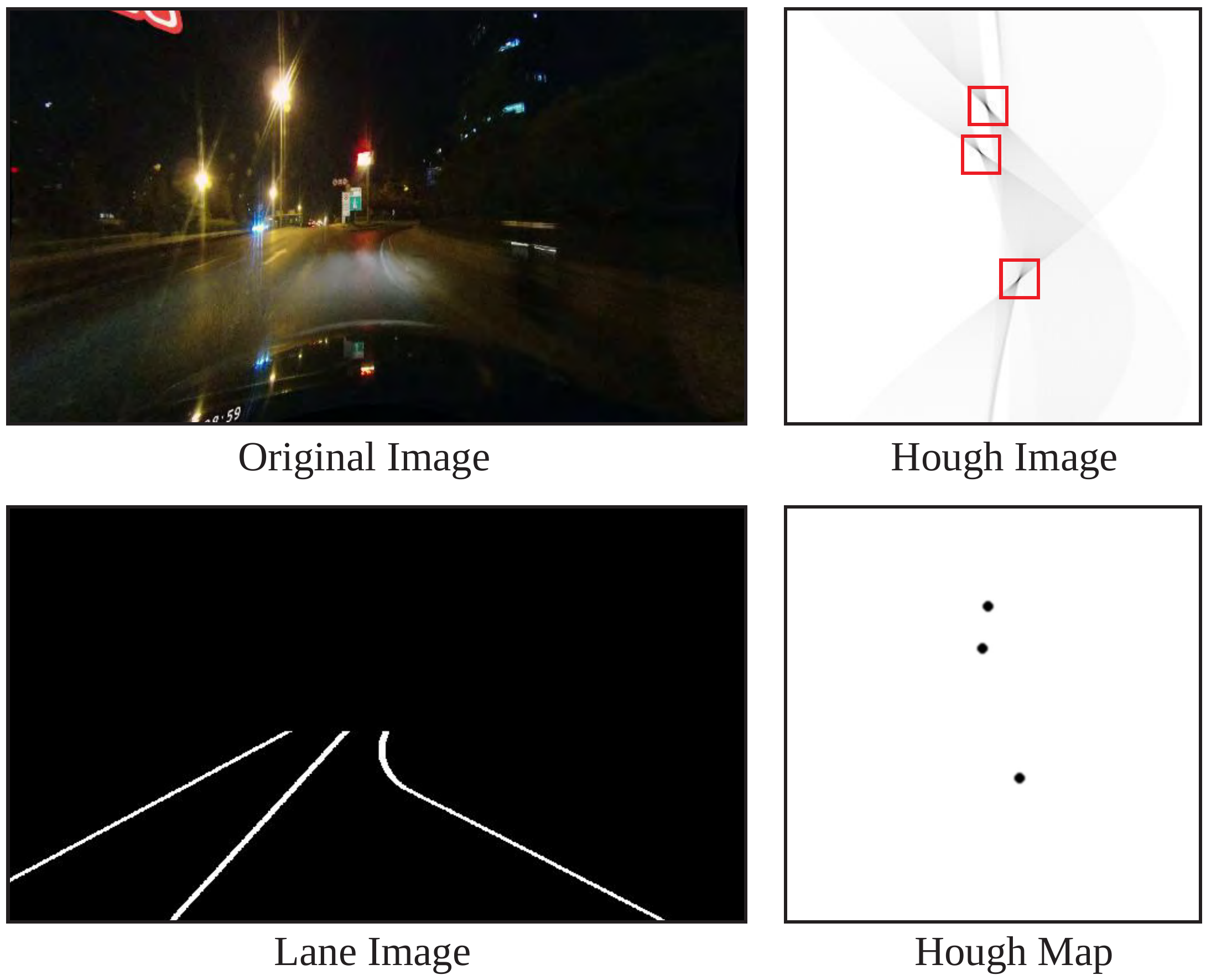}
   \caption{An example of input image and the ground truth lanes from the CULane dataset. The Hough image and Hough map are computed using our method. 
%   Examples of images with lanes and their corresponding Hough image and Hough map in the . 
   In the Hough image, we highlight the area where the Hough points are clustered with a red box.}
   
   \label{fig:hough_curve}
\end{figure}

\textbf{Segmentation-based Methods.\thinspace} 
Segmentation-based methods are one of the most commonly used methods for lane detection ~\cite{offsetmap, hou2020interregion, vangansbeke2019endtoend, SAD, endtoendinstanceseg, SCNN}. They perform binary segmentation for each pixel in the image, where lane pixels are further grouped into lane instances. A segmentation map of each lane is generated through instance segmentation, which is further refined before entering a post-processing step to be decoded into lanes. Pan et al.~\cite{SCNN} treated lane detection as a semantic segmentation task by converting lanes into pixel-level labels. It features a Spatial Convolution Neural Network (SCNN) that performs slice-by-slice convolutions within feature maps, hence enabling message passing between adjacent pixels. Ko et al.~\cite{offsetmap} predicted an offset map for further refinement. Recently, attention modules have been applied to segmentation-based methods. Hou et al.~\cite{SAD} built a soft attention mechanism upon their Self Attention Distillation (SAD) module to generate weight maps for non-significant feature suppression and richer global information distillation. The aforementioned approaches treat lanes as a mere assemblage of individual pixels, disregarding the holistic nature of lanes and neglecting to incorporate the inherent geometry of the lanes. Lin et al.\cite{lin2021semi} employ the line shape prior in their network primarily for the detection of unlabeled lanes in a semi-supervised setting, but with a significantly lower accuracy compared to state-of-the-art methods even with all labels used. Our approach differs from theirs in that we harness the line prior not just for feature aggregation, but also to enhance the process of instance selection.

\begin{figure*}[t!]
   \centering
   \includegraphics[width=0.99\linewidth]{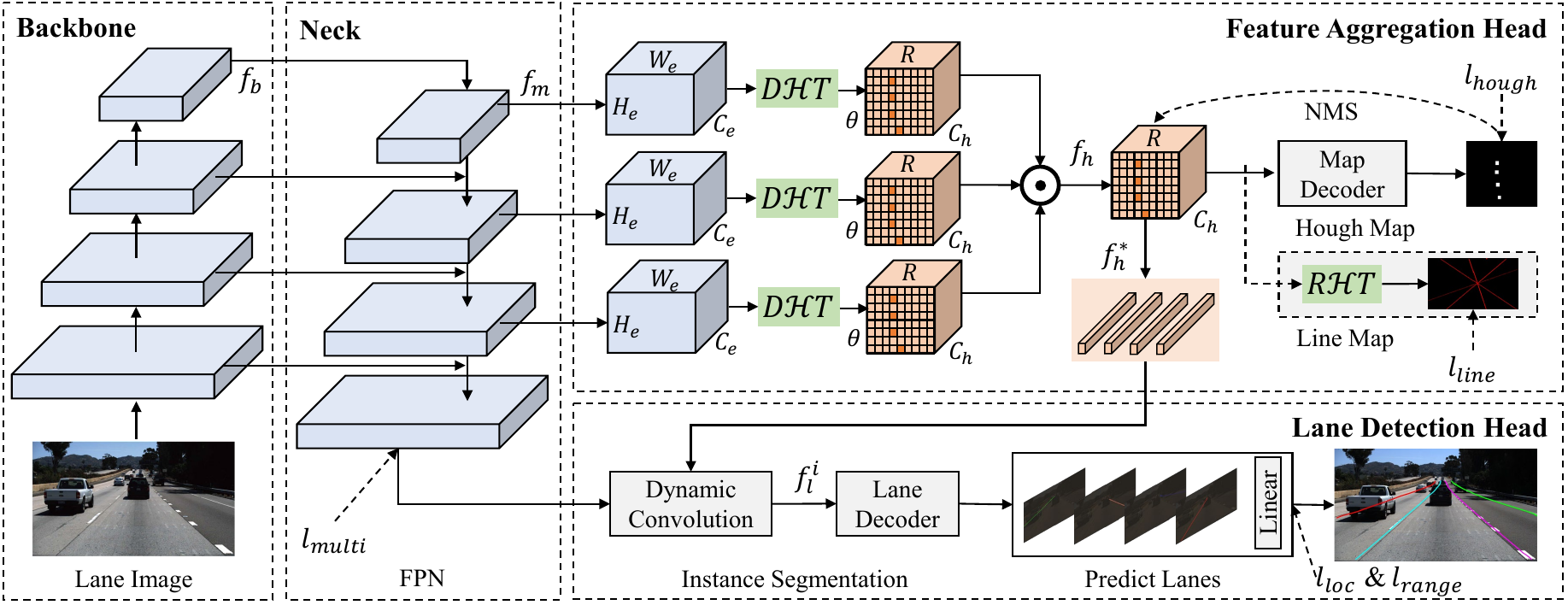}
   \caption{Overview of our HoughLaneNet framework. ResNet~\cite{he2016deep} and FPN~\cite{lin2017feature} are used as the backbone and neck to extract multi-scale features. The Deep Hough Transform (DHT) module transforms Euclidean space features into the parameter space and selects points that correspond to lane features in the parameter space, which are then used to learn the parameters of the dynamic convolution. The lane detection head first inputs Euclidean spatial features into the dynamic convolution to obtain lane maps for different instances, and then utilizes linear units to determine the final lane positions.}
   \label{fig:pipline}
   % \vspace{-1em}
\end{figure*}

\textbf{Anchor-based Methods.\thinspace}
By making use of structural information, Xu et al.~\cite{xu2020curvelanenas} adopted an anchor-based method on the Curvelanes dataset. Each lane can be predicted using vertical anchors and an offset for each point. Tabelini et al.~\cite{LaneATT} improved the representation of anchors %%%built upon their work 
by defining free angle anchors using an origin point $O=(x_{orig},y_{orig})$ on one of the borders of the image (except the top border) and a direction angle $\Theta$. 
%%%\wm{I don't understand the previous sentence.} 
They proposed LaneATT, an anchor-based attention model which makes use of global information to increase the detection accuracy in conditions with occlusion. Su et al.~\cite{structureguided} improved previous techniques by using the vanishing point to guide the generation of lane anchors. Predicted lanes were then generated by predicting an offset from the center line and connecting key points equally spaced vertically. These anchor-based approaches directly inject cognitive bias of common geometry knowledge of lanes into the model, which enables robust prediction on heavily occluded lanes. Nevertheless, the precision still depends on the granularity of the anchor setting.

\textbf{Parametric Methods.\thinspace} 
Tabelini et al.~\cite{tabelini2020polylanenet} proposed to represent lanes as polynomial curves and regress parameters of the curves. Third-order polynomials achieved a competitive accuracy of around 90\% on the TuSimple dataset. As an improvement, Liu et al.~\cite{liu2020endtoend} projected these cubic curves on the ground to the image plane with a new formula and set of parameters and regressed the new curves on the image plane to uniquely identify the lanes. Their end-to-end lane detector built with transformer blocks is able to capture global features to infer lane parameters without being affected by occluded regions. However, those parametric methods can still be inaccurate when generating fine-level lane predictions. \zjq{Fan et al.~\cite{SpinNet_FanWHLM19} introduced an innovative network called SpinNet, which employs a spinning convolution layer to extract lane features at various angles by rotating the feature map. In contrast, our framework uses the Hough transform to gather line features, allowing for the detection of lines at a wider range of angles by increasing the resolution of the Hough map. Furthermore, we can incorporate a Hough map loss function to enhance the learning of network parameters.}

Our work aims to present a segmentation-based method for lane detection that leverages DHT-based feature aggregation to detect lanes with weak visual features. Our method directly isolates each lane feature based on the aggregated Hough features for instance selection, without the need for further multi-class segmentation on the binary-class lane map as required by other segmentation-based methods. Instead,  we only need to predict a binary-class label for each pixel.

\section{Method}

We propose HoughLaneNet, a one-stage lane detection model, and present an overview of its structure in Figure~\ref{fig:pipline}.
% In light of the existing Deep Hough Transform (DHT) module proposed in~\cite{21PAMI-dht}, we propose a one-stage lane detection model, \zjq{HoughLaneNet}. An overview of the \zjq{HoughLaneNet} structure is shown in Figure~\ref{fig:pipline}. 
It receives an input image $I \in \mathbb{R}^{C \times H \times W}$ and applies a backbone (ResNet) to extract deep feature $f_{b}$. Here $C=3$ represents the RGB channels, $H$ and $W$ represent the height and width of the image respectively. Next, we adopt the Feature Pyramid Network (FPN) module or its variations such as PAFPN~\cite{liu2018path} and Recursive-FPN~\cite{qiao2021detectors} to extract multi-scale deep features $f_{m}$. This ensures that our network collects multi-scale information while reducing subsequent computations. 

To deal with the discontinuity, wear and occlusion of lanes, we apply the DHT module to map the multi-scale lane features $f_{m}$ in Euclidean space to $f_{h}$ in the parameter space. We aggregate all features $f_{h}$ in parameter space to serve as the input to the map decoder to obtain the Hough map. This map is used to select Hough features $f^*_h=\{f_{h}^{i}\}_ {i=1}^{N}$ corresponding to the lanes, where $N$ is the number of lanes. The Euclidean multi-scale spatial features $f_{m}$ and selected Hough features $f^*_h$ are fed into the dynamic convolution module to predict deep features of different lanes. Finally, the lane decoder outputs multiple lanes $L=\{l_i\}_{i=1}^N$, where each lane is represented as a series of 2D-points $l_{i}=\{x_k,y_k\}_{k=1}^{H}$.

\subsection{Backbone and Neck}

To extract the backbone deep feature $f_b$, the input image $I \in \mathbb{R}^{C \times H \times W}$ is fed into a backbone network. The backbone network can either be CNN networks such as VGG~\cite{simonyan2014very} or ResNet~\cite{he2016deep}, or attention-based transformer models such as PVT~\cite{wang2021pyramid} or PVTv2~\cite{wang2021pvtv2}. To facilitate comparison with previous methods, we used ResNet pre-trained models as our backbone. In terms of network parameters, ResNet18, ResNet34, ResNet101 are used as the small, medium and large versions of HoughLaneNet, respectively.

In the object detection task, extracting multi-scale deep features plays a key role in performance improvement. There are usually two structures of FPN and ASPP to help the network extract multi-scale information. In our case, we used FPN to extract multi-scale features $f_m$. During the early stages of training, the backbone may not accurately recognize which regions in the image correspond to lanes, leading to an unstable prediction of the Hough map by the DHT module. To mitigate this, we impose constraints on the multi-scale feature $f_m$ output generated by the neck, thereby enabling the backbone to quickly distinguish the pixels in the image that correspond to lanes. We decode $f_m$ to a binary image $\hat{y}_{b}$ by multi-layer convolution, and binary-cross-entropy loss is used to guide the network training:

\begin{equation}
l_{multi}=-(y_{b} \times log(\hat{y}_{b}) + (1-y_{b}) \times log(1-\hat{y}_{b})),
\end{equation}
\noindent
where $y_b$ represents the ground truth lane probability map, and $\hat{y}_{b}$ is the predicted lane probability map. In Section~\ref{sec:ablation} we provide experiments for the influence of different pre-trained models as the backbone, and various versions of FPN as the neck, on the detection results.

\begin{figure}[t!]
   \centering
   \includegraphics[width=0.9\linewidth]{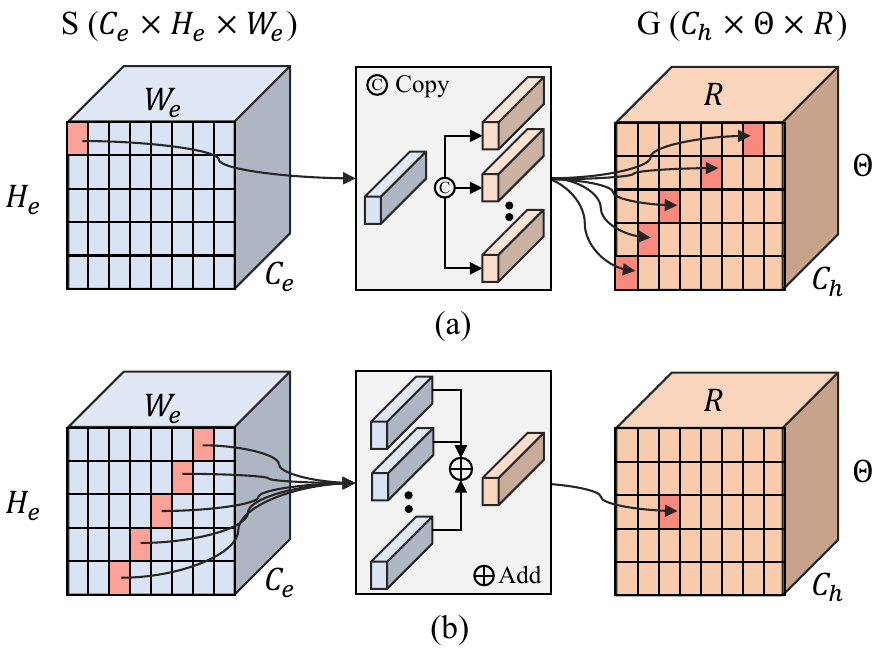}
   \caption{The process of Deep Hough Transform. (a) illustrates that the feature of a point ($h$, $w$) in $S$ is copied and added to the corresponding positions in $G$. (b) illustrates that after transformation, features belonging to the same line are clustered to the same position in the parameter space.
   %%% \wm{$\Theta$ or $\theta$? Caption (a) and (b) should use a smaller font.}
   }
   \label{fig:hough_trans}
\end{figure}

\subsection{Feature Aggregation}

To transform the multi-scale features $f_m$ into the lane parameter space, we first interpolate the FPN outputs to the same size and then apply DHT, which aggregates the local features of all points on the lane with visual features. The multi-scale deep features reflect the position of the lanes in Euclidean space. After transforming to the parameter space, all deep features belonging to the same straight line will be clustered to the same position in the space. This enhances the features of discontinuous, worn, and occluded lanes which would otherwise be very difficult to detect. In this section, we will first introduce the process of DHT, and then describe how HoughLaneNet uses DHT to find the location of lanes.

\subsubsection{Deep Hough Transform}

A point ($\theta,r$) in the polar coordinate system can be used to represent a straight line in a 2D image using Hough transform according to the following transformation formula:

\begin{equation}\label{eq:hough_trans}
r = x\cos\theta + y\sin\theta.
\end{equation}

Following DHT-based line detection~\cite{21PAMI-dht}, we define $ \Theta $ and $R$ as the number of quantization levels of $\theta$ and $r$, respectively, where $\Theta = 360$ and $R = 360$. As shown in Figure~\ref{fig:hough_trans}, the spatial feature $S \in (C_e \times H_e \times W_e)$ represents the image features in the Euclidean space, and the Hough feature $G \in (C_h \times \Theta \times R)$ represents the features in parameter space.

Figure~\ref{fig:hough_trans} (a) shows how each point feature in Euclidean space is transformed into parameter space. Given a point $(x, y)$ in the Euclidean space $S$, by traversing all $\Theta$ values, we can get a series of points $\{\theta_{i}, r_{i}\}_{i=1}^{\Theta}$ in the parameter space $G$ through Equation~\ref{eq:hough_trans}. The DHT will copy the feature of point $(x, y)$ to each point in the polar coordinate system.

Since points on the same line in Euclidean space $S$ are transformed to the same position in parameter space, the features of each point on the line in Euclidean space are added together, forming a point feature in the parameter space $G$, as shown in Figure~\ref{fig:hough_trans} (b). In practice, $G$ is initialized with zero padding, and then a DHT is applied to each point in $S$.

\begin{figure}[t!]
   \centering
   \includegraphics[width=0.95\linewidth]{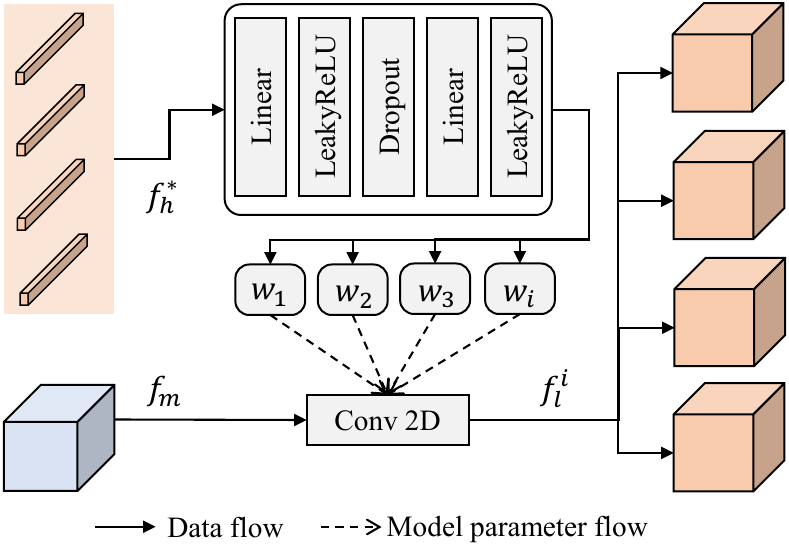}
   \caption{The dynamic convolution module. Input the Hough features $f^*_h$ and multi-scale features $f_m$, and output the instance features of different lanes $f_{l}^{i}$. It is noteworthy that our model possesses the capability to predict a variable number of lanes, despite the illustration in the figure only depicting four lanes.}
   \label{fig:dynamic_conv}
\end{figure}

\subsubsection{Lane Feature Aggregation}

Our feature aggregation head mainly consists of DHT modules from three layers, as is shown in Figure~\ref{fig:pipline}. The DHT modules are used to aggregate features $f_m$ of different lanes in different scales, which helps to amplify the features of the lanes. Moreover, in an attempt to save computational resources, we only transform three layers of the high-dimensional features and discard the first layer of features from the FPN output features. The transformed three layers of Hough features at different scales are first up-sampled to a uniform size and concatenated together, then passed through a map decoder to decode the transformed features $f_h$ to create the Hough map, which is the same resolution as the ground truth Hough map. We observe that the number of positive samples in the Hough map is much less than the number of negative samples, therefore the focal loss is used to constrain the Hough map in order to improve the problem of data sample imbalance following CondLaneNet~\cite{condlanenet} and CenterNet~\cite{duan2019centernet}:

\begin{equation}  
l_{hough}=\frac{-1}{N}\sum_{\theta r}
\left\{ 
    \begin{array}{lc}
        (1-\hat{P}_{\theta r})^{\alpha}log(\hat{P}_{\theta r}) & P_{\theta r} = 1 \\
        (1-P_{\theta r})^{\beta}(\hat{P}_{\theta r})^{\alpha}log(1-\hat{P}_{\theta r}) & P_{\theta r} \neq 1 \\
    \end{array}
    ,
\right.
\end{equation}

\noindent
where $N$ is the number of lanes in the input image, and we follow CondLaneNet to set $\alpha$ and $\beta$ to $2$ and $4$ respectively. $\hat{P}_{\theta r}$ is the predicted probability at point $(\theta,r)$, and $P_{\theta r}$ is the ground truth probability at point $(\theta,r)$ of the Hough map.

In our experiments we notice that the loss added on the Hough map does not give an accurate position, and the actual predicted position of the Hough point hovers around the ground truth. Therefore, we use reverse Hough transform (RHT) to transform $f_h$ to Euclidean space again, and add a loss term to predict the line map. Here, the line map is not the position of lanes, but is instead a transformation of each point in the Hough map to a straight line. We use binary-cross-entropy loss $l_{line}$ to constrain the line map:

\begin{equation}
l_{line}=-(y_{b} \times log(\hat{y}_{b}) + (1-y_{b}) \times log(1-\hat{y}_{b})),
\end{equation}

\noindent
where $y_b$ represents the ground truth line probability map, and $\hat{y}_{b}$ is the predicted line probability map. Please refer to Section~\ref{sec:details} for the details of aggregating each lane to a point in Hough map.

\subsubsection{Hough Feature Selection}\label{sec:hough_select}

% Each point of the Hough feature $f_h$ in the %%% two-dimensional 
% \zjq{parameter} space represents the global feature of each lane. We locate the Hough feature $f_h^i$ of each lane by counting the number and finding the positions of the points on the Hough map. \wm{what does `counting the number and finding the positions of the points on the Hough map' mean? How you did it?} Subsequently, we treat$f^*_h=\{f_{h}^{i}\}_ {i=1}^{N}$ as learned parameters of the dynamic convolution to segment lanes, as described in \cref{sec:instance}. In the training phase, ground truth points of the Hough map are given directly and compared against. However during test time, non-maximum suppression (NMS) procedure is used to extract the locations of Hough points from the Hough map. Specifically, the Hough map is fed into a max-pooling layer with stride $1$, and all non-max pixels in the Hough map are set to zero. Finally, the Hough points are determined by threshold filtering \wm{should provide details of threshold filtering???}. 

Each point of the Hough feature $f_h$ in the parameter space represents the global feature of each lane. We locate the Hough feature $f_h^i$ of each lane by applying a non-maximum suppression (NMS) procedure. Specifically, the Hough map is fed into a max-pooling layer with stride $1$ and kernel size $5$, and all non-max pixels in the Hough map are set to zero. Finally, the Hough points are determined by threshold filtering. Compared to other NMS calculation methods, the aforementioned strategy yields higher FPS. In supplementary material, %section~\ref{sec:ablation_dht}, 
we display and analyze the impact of various point selection strategies on the results. In practice, we set different thresholds for the three test datasets, $0.1$ for Tusimple dataset and $0.15$ for CULane and LLAMAS. In the proposed method, the NMS procedure is only applied during the test phase, and ground truth points of the Hough map are given directly and compared against in the training phase. Subsequently, we treat $f^*_h=\{f_{h}^{i}\}_ {i=1}^{N}$ as learned parameters of the dynamic convolution module to segment lanes, as described in Section~\ref{sec:instance}.

\subsection{Lane Detection Head}\label{sec:instance}

With all the selected Hough features $f^*_h$, the lane detection head mainly applies a dynamic convolution module for instance detection, as is shown in Figure~\ref{fig:dynamic_conv}. The dynamic convolution contains one layer of 2D convolution, and its kernel parameters are mainly obtained by the selected Hough features $f^*_h$ which are fed into a multi-layer perception (MLP) to regress the parameters. The multi-scale features $f_m$ are passed through a dynamic convolution of different kernel parameters, which outputs features of different lanes $\{f_{l}^{i}\}_ {i=1}^{N}$ and feeds them into a lane decoder to obtain location maps of lanes, and finally the positions of lane points are predicted by the location map.

\begin{figure}[t!]
   \centering
   \includegraphics[width=0.9\linewidth]{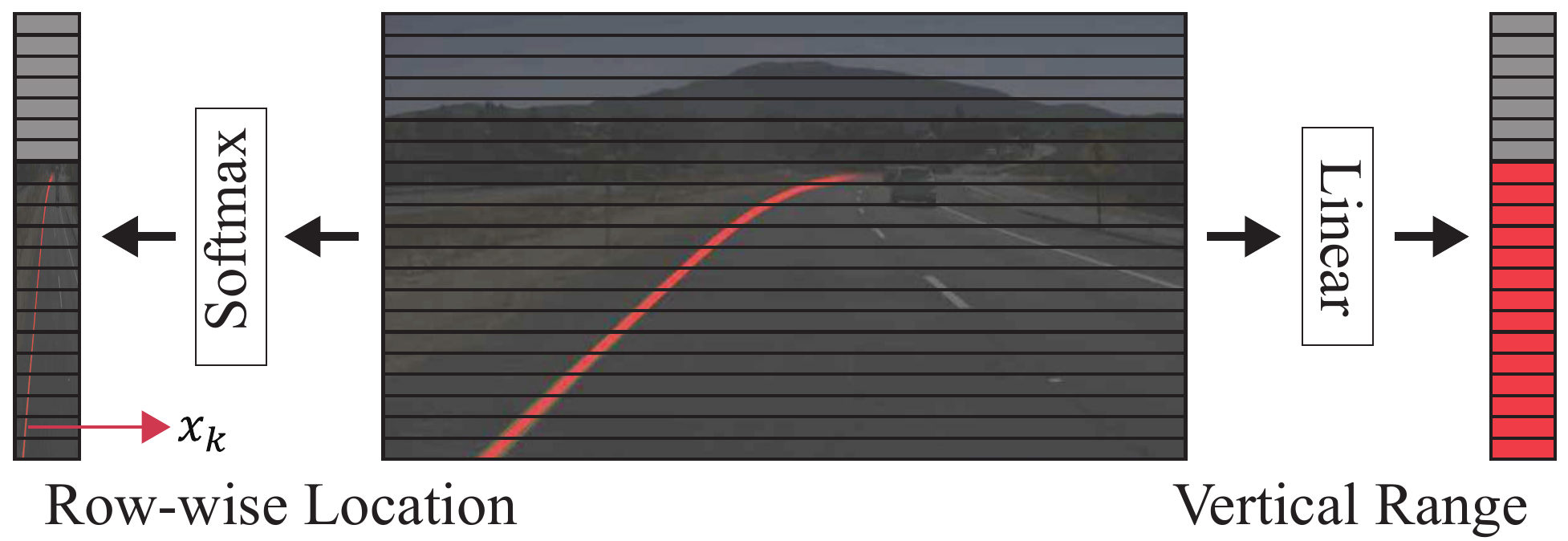}
   \caption{The process of predicting lane row location from the location map. Each row of the location map is first passed through the linear layer to determine which row the lane passes through. Rows identified with lanes are then passed through a softmax layer to obtain the row location.}
   \label{fig:lane_format}
\end{figure}

For the location map of each lane, we use the position-weighted binary cross-entropy loss function to constrain the predicted location maps, and all location maps are concatenated together when calculating the loss:

\begin{equation}
l_{loc}=-\sum_{c=1}^{N}p_c \times y_{bc} \times log(\hat{y}_{bc}) + (1-y_{bc}) \times log(1-\hat{y}_{bc}),
\end{equation} 

\noindent
where $y_{bc}$ is the ground truth lane probability map, $\hat{y}_{bc}$ is the predicted lane probability map and $p_{c}$ represents the weight of the positive class. $N$ is the number of classes, which is equal to the number of lanes. 

Since the $y$-coordinate of each lane $\{y_k\}_{k=1}^{H}$ can be determined given the vertical range and image height $H$, we mainly focus on the $x$-coordinate, $\{x_k\}_{k=1}^{H}$ of each row. Our row-wise formulation is different from CondLaneNet~\cite{condlanenet} and Ultrafast~\cite{ultrafast}, in the sense that we require only two quantities: the predicted vertical range and row-wise location. Figure~\ref{fig:lane_format} shows a predicted lane location map, which is an image of size $H \times W$. To obtain the $x$-coordinate of each lane, each row in the location map is first fed into the linear layer to obtain a binary classification result. This classification determines which rows have a lane passing through it. Subsequently, the feature vector of the row with the lane passing through is sent to the softmax layer to obtain the final $x$-coordinate. Rows in the location map with no lanes passing through them, are filled with default values.

Following CondLaneNet we use softmax-cross-entropy $l_{range}$ to guide the network to learn the vertical range. Combining all of the above loss terms together, our overall training loss is defined as follows:

\begin{equation}
l_{total}=\lambda_{m}l_{multi} + \lambda_{h}l_{hough} + \lambda_{l}l_{line} + \lambda_{c}l_{loc} + \lambda_{r}l_{range},
\end{equation}

\noindent
where $\lambda$ values are different weights added to each term. In practice, we found that the computed loss values are very small, especially the loss value constraining the Hough map. Therefore, we set different loss weights based on experimental experience and considering the importance of different loss terms. Specifically, $\lambda_{m}=100$, $\lambda_{h}=1000$, $\lambda_{l}=100$, $\lambda_{c}=100$ and $\lambda_{r}=10$.

\section{Experiments}

We evaluate HoughLaneNet on the TuSimple~\cite{TuSimple}, CULane~\cite{SCNN} and LLAMAS~\cite{llamas2019} datasets, and demonstrate our method through quantitative and qualitative comparisons. In the following sections, we first introduce the processing of datasets and the experimental environment, then show the comparison results with the current state-of-the-art methods, and finally analyze the effectiveness of the network components on the results.

\subsection{Datasets and Metrics}
% \label{sec:datasets}

The TuSimple dataset contains $6408$ annotated high-resolution ($1280 \times 720$) images taken from video footage, each of $20$ frames shot on highway lanes in the U.S. with light traffic. Most clips in this dataset are taken during daytime under good weather conditions, with $2$ to $4$ lanes with clear markings in each image. The clear lane markers greatly simplify the detection task, and classic segmentation-based methods can achieve high accuracy without the need for heavy enhancements. Previous state-of-the-art algorithms which do not require extra training data achieved an extremely high accuracy of $96.92\%$~\cite{folo} on TuSimple. 

The CULane dataset contains clips from a video of more than $55$ hours in length, which was collected by drivers in Beijing city streets. Due to the crowdedness and complexity of city streets, lane markings are often either occluded or unclear due to abrasion, which makes detection much harder without a complete understanding of the shape of lanes and the semantics of the scene. The test set in CULane is divided into normal and eight challenging categories.

The LLAMAS dataset contains more than 100,000 high-resolution images ($1276 \times 717$), collected primarily from highway scenarios. It is automatically annotated for lane markers using Lidar maps. Following existing lane detection algorithms~\cite{LaneATT} on the LLAMAS dataset, we detect only four main lanes from images. Since there are no public annotations of the test set, we upload the test results to the official website to get the evaluation results. Details of the three datasets are shown in Table~\ref{tab:info_datasets}.

\begin{figure}[t!]
   \centering
   \subfigure[]{\includegraphics[width=0.48\linewidth]{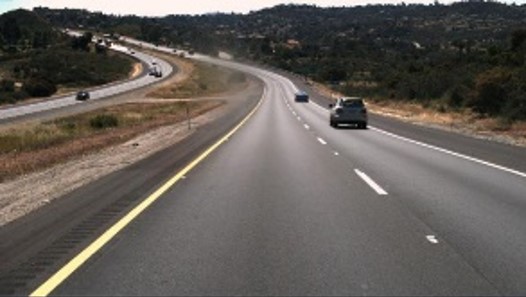}}
   \subfigure[]{\includegraphics[width=0.48\linewidth]{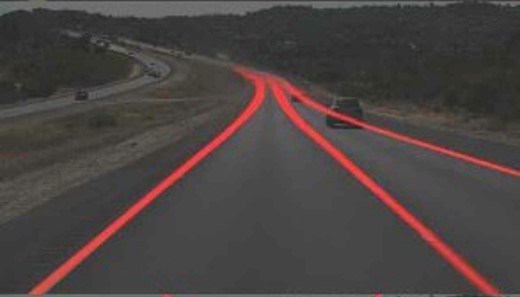}}\\
   
   \subfigure[]{\includegraphics[width=0.27\linewidth]{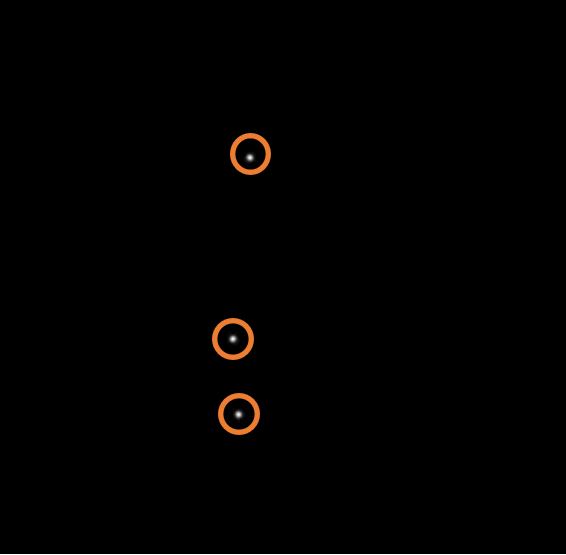}}
   \subfigure[]{\includegraphics[width=0.48\linewidth]{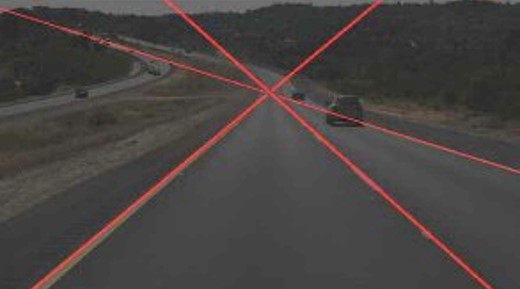}}
   \caption{An example visualization of Hough map construction. (a) is the input image, (b) shows the ground truth lanes, the points in the orange circle in (c) are constructed Hough points, and (d) shows lines that correspond to the reversed Hough points from (c) in Euclidean space.}
   \label{fig:hough_point}
\end{figure}

\begin{table}[t!]
    \begin{center}
    \caption{Details of the three datasets for evaluation.\\}
      \resizebox{\columnwidth}{!}{%
    \begin{tabular}{lllll}
    % \hline
    % Dataset    & Train & Val. & Test  & Resolution  & Scenarios \\ 
    % \hline
    % TuSimple   & 3.3K  & 0.4K & 2.8K & ($1280, 720$) & Highway \\
    % CULane    & 88.9K & 9.7K & 34.7K & ($1640, 590$) & Urban+Highway \\
    % LLAMAS    & 58.3K & 20.8K & 20.9K & ($1276, 717$) & Highway \\

    \hline
    Dataset    & Train & Test  & Resolution  & Scenarios \\ 
    \hline
    TuSimple   & 3.7K & 2.8K & ($1280, 720$) & Highway \\
    CULane    & 98.6K & 34.7K & ($1640, 590$) & Urban+Highway \\
    LLAMAS    & 79.1K & 20.9K & ($1276, 717$) & Highway \\

    \hline
    \end{tabular}
    }
        \label{tab:info_datasets}
    \end{center}

\end{table}

We report both the accuracy and F1-score on all datasets. The official indicator in the TuSimple dataset includes only accuracy, but we added the F1-score following~\cite{LaneATT, condlanenet}. The CULane and LLAMAS datasets are evaluated by official evaluation indicators~\cite{SCNN} which uses the F1-measure as the metric. The accuracy is calculated as:
\begin{equation}
Accuracy = \frac{N_{pred}}{N_{gt}},
\end{equation}
where $N_{pred}$ is the number of lane points that have been correctly predicted and $N_{gt}$ is the number of ground-truth lane points.
We also report the F1-measure, which is based on the intersection over union (IoU) of predicted lanes and ground truth lanes. Lanes that have IoU greater than $0.5$ are considered correct predictions. For the CULane dataset, only the F1-score is reported. This is calculated as:
\begin{equation}
F1-measure = 2 \cdot \left( \frac{precision \cdot recall}{precision + recall} \right),
\end{equation}
\begin{equation}
precision = \frac{TP}{TP+FP},
\end{equation}
\begin{equation}
recall = \frac{TP}{TP+FN}, 
\end{equation}
where $TP$ is the number of lane points correctly predicted, $FP$ and $FN$ are the numbers of false positive and false negative lane points respectively.

\begin{figure}[t!]
   \centering
   \subfigure[]{\includegraphics[width=0.32\linewidth]{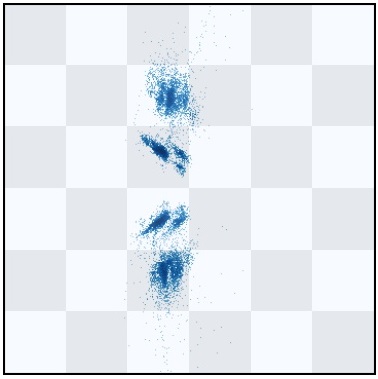}}
   \subfigure[]{\includegraphics[width=0.32\linewidth]{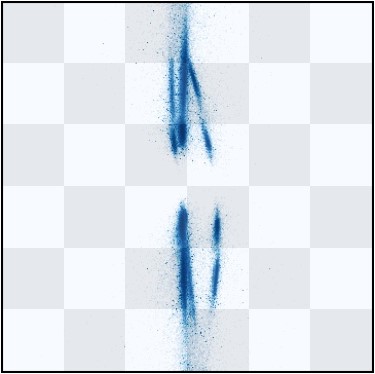}}
   \subfigure[]{\includegraphics[width=0.32\linewidth]{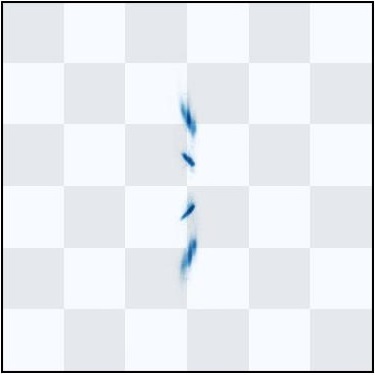}} \\
   \subfigure[]{\includegraphics[width=0.32\linewidth]{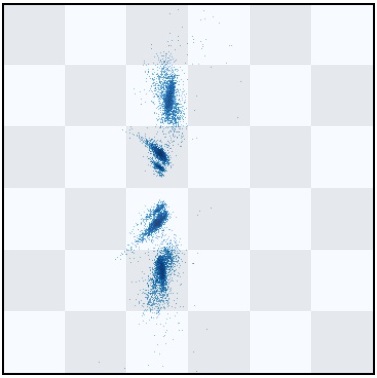}}
   \subfigure[]{\includegraphics[width=0.32\linewidth]{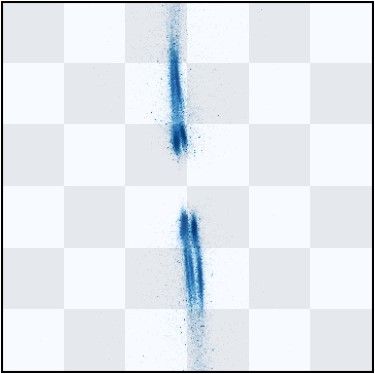}}
   \subfigure[]{\includegraphics[width=0.32\linewidth]{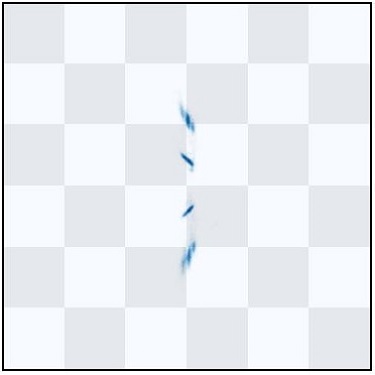}}
   \caption{Distribution of lanes in parameter space. (a), (b) and (c) provide a visualization of lanes in parameter space from the TuSimple, CULane and LLAMAS train dataset respectively. (d) and (e) are the visualizations of lanes in parameter space from the TuSimple and CULane test dataset respectively. (f) shows the visualization of lanes from the LLAMAS validation dataset.}
   \label{fig:hough_dist}
\end{figure}

\begin{figure*}[ht!]
   \centering
   \includegraphics[width=0.99\linewidth]{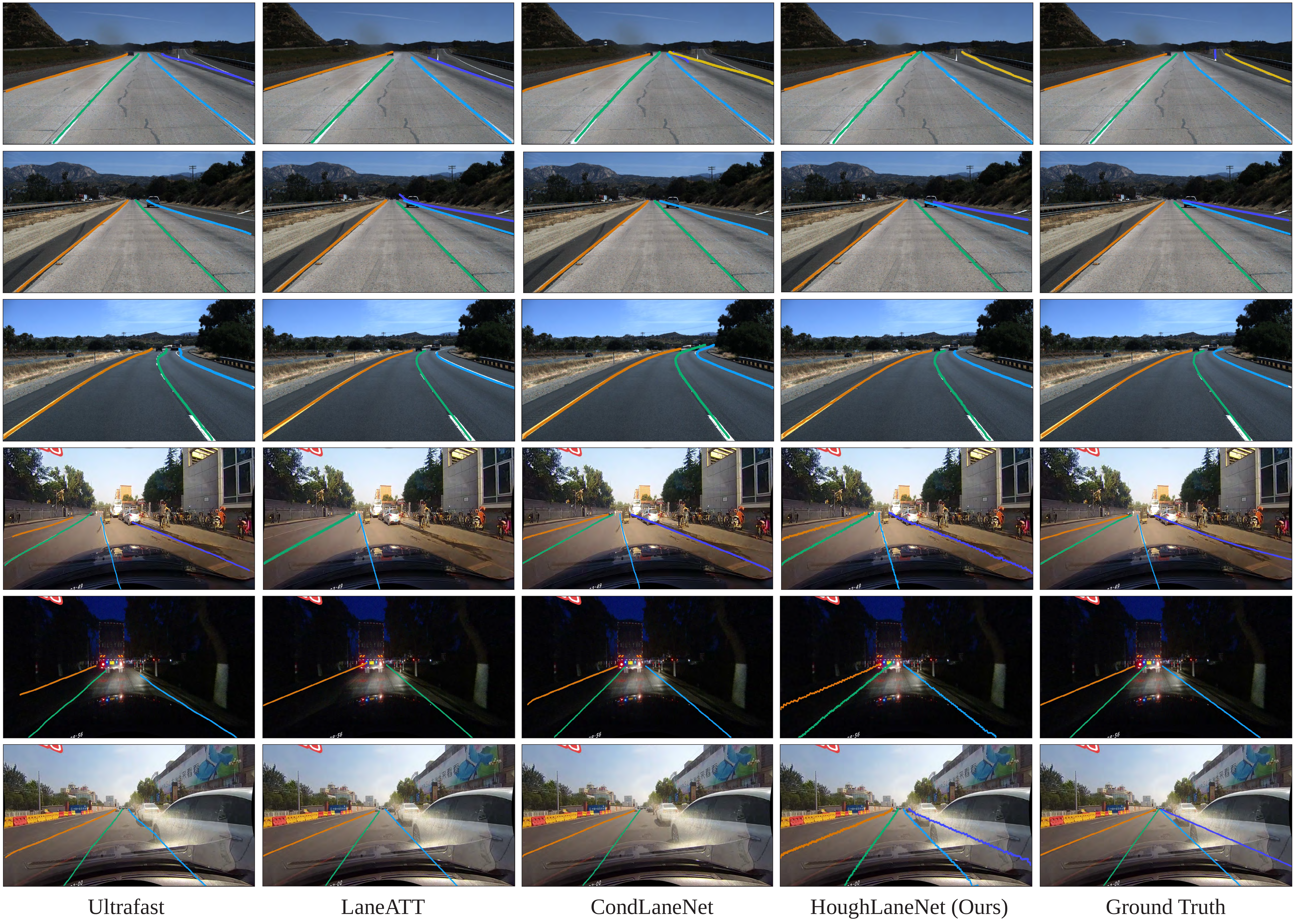}
   \caption{Visualization of lane detection results for different models on the TuSimple and CULane dataset. The first, second and third rows show results on the TuSimple dataset. The remaining rows show results on the CULane dataset. The detection results of different methods (Ultrafast~\cite{ultrafast}, LaneATT~\cite{LaneATT} and CondLaneNet~\cite{condlanenet}) are displayed in turn from left to right, and the fourth column shows the results of our proposed HoughLaneNet.}
   \label{fig:example_lane}
   % \vspace{-1em}
\end{figure*}

\begin{table}[t!]
\setlength\tabcolsep{4pt}
\begin{center}
\caption{Different configurations of HoughLaneNet for the TuSimple dataset.\\}
  \resizebox{\columnwidth}{!}{%
\begin{tabular}{lcccc}
\hline
Config & \begin{tabular}[c]{@{}c@{}}Backbone\\ Net\end{tabular} & \begin{tabular}[c]{@{}c@{}}Hough Map\\ Size\end{tabular} & \begin{tabular}[c]{@{}c@{}}Hough Head\\ Channels\end{tabular} & \begin{tabular}[c]{@{}c@{}}Instance\\ Channels\end{tabular} \\ \hline
Small  & ResNet18   & (240,240)    & 128    & 32  \\
Medium & ResNet34   & (300,300)    & 128    & 32  \\
Large  & ResNet101   & (360,360)    & 192    & 48  \\ \hline
\end{tabular}
}
\label{table:config}
\end{center}

\end{table}

\subsection{Implementation Details}\label{sec:details}

A key problem for our training process is how to obtain the ground truth Hough label maps of the lanes. Considering that lanes are mostly straight at the bottom of the image and may be curved at the top of the image, we sample a fixed number of lane points with an equal interval at the bottom of the image. Next, we use pairs of adjacent points to define straight lines and map them to the corresponding point ($\theta,r$) in parameter space. Finally, the average coordinates of all such points in parameter space is defined as the point of this lane, as illustrated in Figure~\ref{fig:hough_point}. 
%It can be seen from (b) and (d) in Figure~\ref{fig:hough_point} that our method can effectively construct accurate Hough points.

We present a distribution anlysis of constructed Hough labels for the training and test set on the TuSimple, CULane and LLAMAS, as shown in Figure~\ref{fig:hough_dist}. Horizontal and vertical axes denote $R$ and $\Theta$ respectively. The distribution of lane points on the TuSimple and LLAMAS datasets exhibit a more uniform pattern and are easier to learn compared to the CULane dataset. Specifically, the lanes in TuSimple and LLAMAS datasets distribute across four distinct regions with defined boundaries, while in the CULane dataset, the lane lines occupy approximately $80\%$ of the angle. Therefore, We choose different sized Hough map for each dataset. Additionally, it is worth mentioning that while the majority of lanes in the dataset fall within a narrow range of angles, there are some outlier lanes that are distributed across the entire range of angles. For further discussion on the impact of Hough map size on results, see supplementary material.
% Sec~\ref{sec:ablation_dht}.} 

We provide three configurations of the network for different computational requirements: Small, Medium and Large. The configurations differ in the backbone net, the size of Hough map, the number of channels in the Hough proposal head and the instance head for TuSimple dataset, as shown in Table~\ref{table:config}. For the CULane and LLAMAS datasets, the configurations in Table~\ref{table:config} are also applicable, except that the Hough map size is changed: (360,216) for the CULane dataset, and (360,360) for the LLAMAS dataset. The resolution of Hough features $f_h$ is set to $1/3$ of the ground-truth Hough map resolution for all configurations and datasets. Different threshold values for NMS instance proposal procedure are used for each dataset to achieve best result in the validation step, see Section~\ref{sec:hough_select}. Our backbone network does not take the raw image as input, but rather a down-sampled image with a resolution of $640 \times 360$ as an input in both the training and testing stage. During training, the raw images are augmented with RGB shift, hue saturation, blurring, color jittering and random brightness.

We use the AdamW optimizer with a learning rate first set to $3e^{-4}$ with a step decay learning rate strategy and constant warm-up policy, which subsequently drops by a ratio of $0.9$ for every $15$ epochs in TuSimple and each epoch in CULane and LLAMAS. We train $70$ epochs on TuSimple, $20$ epochs on CULane and $18$ epochs on LLAMAS with a batch size of $3$, $2$, and $2$ for the Small, Medium, and Large configurations respectively. We conduct our experiments with three RTX3090 GPUs.

% \wm{I am here.} 

\begin{table}[t!]
    \centering
        \captionof{table}{Comparison of the performance of different methods on TuSimple. The best results for each metric are given in bold, and F1 scores are computed using the official code.\\}
    \begin{tabular}{lcccccc}
    \hline

        \textbf{Method}         & \textbf{Backbone}            & \textbf{F1$\uparrow$} & \textbf{Acc$\uparrow$} & \textbf{FPS$\uparrow$} \\ \hline
    Eigenlanes~\cite{jin2022eigenlanes} & \multicolumn{1}{c}{-} & \multicolumn{1}{c}{-}  & 95.62  & \multicolumn{1}{c}{-} \\ 
    SCNN~\cite{SCNN} & VGG16 & 95.97  & 96.53   & 7.5  \\
    ENet-SAD~\cite{SAD}  & \multicolumn{1}{c}{-} & 95.92  & 96.64  & 75.0 \\ 
    RESA~\cite{zheng2021resa} & ResNet34 & \multicolumn{1}{c}{-} & 96.82  & \multicolumn{1}{c}{-} \\
    LaneAF~\cite{abualsaud2021laneaf}  & DLA-34 & 96.49  & 95.62  & \multicolumn{1}{c}{-} \\ 
    LSTR~\cite{liu2020endtoend}      & ResNet18              & \multicolumn{1}{c}{96.86} & 96.18  & \textbf{420}   \\
    
    LaneATT~\cite{LaneATT} & ResNet18 & 96.71                 & 95.57   & 250               \\
    LaneATT~\cite{LaneATT} & ResNet34 & 96.77                 & 95.63   & 171               \\
    LaneATT~\cite{LaneATT} & ResNet122 & 96.06                 & 96.10   & 26        \\
    CondLane~\cite{condlanenet} & ResNet18 & 97.01                 & 95.48   & 220       \\
    CondLane~\cite{condlanenet} & ResNet34 & 96.98                 & 95.37   & 154           \\
    CondLane~\cite{condlanenet} & ResNet101 & 97.24    & 96.54   & 58                  \\
    FOLOLane~\cite{folo}  & ERFNet & 96.59 & 96.92 & \multicolumn{1}{c}{-} \\
    CLRNet~\cite{zheng2022clrnet} & ResNet18 & \textbf{97.89}                 & 96.84  & \multicolumn{1}{c}{-}     \\
    CLRNet~\cite{zheng2022clrnet} & ResNet34 & 97.82                 & 96.87  & \multicolumn{1}{c}{-}           \\
    CLRNet~\cite{zheng2022clrnet} & ResNet101 & 97.62    & 96.83   & \multicolumn{1}{c}{-}            \\
    \hline
    %PolyLaneNet             & 90.62  & 93.36   & 9.42    & 9.33    & 115.0 & 1.7     \\ \hline
    % \textbf{HoughLane}            &  ResNet18       &   97.46      &    97.00     &     3.00    &   2.07    &    100     \\
    
    % \textbf{HoughLane}            &  ResNet18       &   97.74      &    96.70     &     2.10    &   2.41    &    131     \\
    % % epoch=140 | th=0.1 | point 
    \textbf{HoughLane}            &  ResNet18       & 97.67  & 96.71  & 172  \\
    % epoch=140 | th=0.1 | NMS 
    \textbf{HoughLane}            &  ResNet34      &    97.68     &     \textbf{96.93}   &    125     \\
    % epoch=145 | th=0.1 | NMS
    \textbf{HoughLane}            & ResNet101       &  97.31 & 96.43  & 57    \\
    % epoch=136 | th=0.1 | nms

    \hline
    \end{tabular}
    \label{table:test_tusimple}
\end{table}

\begin{table*}[t!]
    \setlength\tabcolsep{4pt}
    \centering
    \caption{Comparison of the performance of different methods on CULane. Here, the total score is calculated as the overall F1 score on the test set. For the ``Cross" category, the number of false positives is reported as no lanes are on the image. The best results for each metric are given in bold. Results are tested with FPN, while other necks can also be used.\\}
  \resizebox{\linewidth}{!}{%
    \begin{tabular}{lcccccccccccc}
    \hline
    \textbf{Method}     & \textbf{Backbone}           & \textbf{Total$\uparrow$} & \textbf{Normal}      & \textbf{Crowded} & \textbf{Dazzle}      & \textbf{Shadow} & \textbf{No line}     & \textbf{Arrow}       & \textbf{Curve}          & \textbf{Cross} & \textbf{Night} & \textbf{FPS$\uparrow$}   \\ \hline
    ERFNet-E2E~\cite{endtoend2020} & - & 74.00 & 91.00       & 73.10   & 64.50       & 74.10  & 46.60       & 85.80       & 71.90          & 2022  & 67.90 &  \\
    SCNN~\cite{SCNN} & VGG16  & 71.60 & 90.60       & 69.70   & 58.50       & 66.90  & 43.40       & 84.10       & 64.40          & 1990  & 66.10 & 7.5      \\
    ENet-SAD~\cite{SAD} & - & 70.80 & 90.10       & 68.80   & 60.20       & 65.90  & 41.60       & 84.00       & 65.70          & 1998  & 66.00 & 75            \\
    Ultrafast~\cite{ultrafast} & ResNet18 & 68.40 & 87.70       & 66.00   & 58.40       & 62.80  & 40.20       & 81.00       & 57.90          & 1743  & 62.10 & \textbf{322.5}   \\
    Ultrafast~\cite{ultrafast}   & ResNet34 & 72.30 & 90.70       & 70.20   & 59.5       & 69.30  & 44.40       & 85.70       & 69.50          & 2037  & 66.70 & 175.4   \\
    CNAS-S~\cite{xu2020curvelanenas} & -  & 71.40 & 88.30       & 68.60   & 63.20       & 68.00  & 47.90       & 82.50       & 66.00          & 2817  & 66.20 &   \\
    CNAS-M~\cite{xu2020curvelanenas} & - & 73.50 & 90.20       & 70.50   & 65.90       & 69.30  & 48.80       & 85.70       & 67.50          & 2359  & 68.20 &               \\
    CNAS-L~\cite{xu2020curvelanenas} & - & 74.80 & 90.70       & 72.30   & {67.70} & 70.10  & {49.40} & 85.80       & {68.40}    & 1746  & 68.90 &             \\ 
    LaneAF~\cite{abualsaud2021laneaf} &  DLA-34       & 77.41 & 91.80       & 75.61   & 71.78 & 79.12  & 51.38       & 86.88       & 72.70          & 1360  & 73.03 &           \\
    SGNet~\cite{structureguided} & ResNet34 & 77.27 & 92.07       & 75.41   & 67.75       & 74.31  & 50.90       & 87.97       & 69.65          & 1373  & 72.69 & 92            \\
    LaneATT~\cite{LaneATT} & ResNet18 & 75.13 & 91.17       & 72.71   & 65.82       & 68.03  & 49.13       & 87.82       & 63.75          & \textbf{1020}  & 68.58 & 250                \\
    LaneATT~\cite{LaneATT} & ResNet34 & 76.68 & 92.14       & 75.03   & 66.47       & 78.15  & 49.39       & 88.38       & 67.72          & 1330  & 70.72 & 171           \\
    LaneATT~\cite{LaneATT} & ResNet122 & 77.02 & 91.74       & 76.16   & 69.47       & 76.31  & 50.46       & 86.29       & 64.05          & 1264  & 70.81 & 26        \\ 
    CondLaneNet~\cite{condlanenet} & ResNet18 & 78.14 & 92.87       & 75.79   & 70.72       & 80.01  & 52.39       & 89.37       & 72.40          & 1364  & 73.23 & 220           \\
    CondLaneNet~\cite{condlanenet} & ResNet34 & 78.74 & 93.38       & 77.14   & 71.17       & 79.93  & 51.85       & 89.89       & 73.88          & 1387  & 73.92 & 152           \\
    CondLaneNet~\cite{condlanenet} & ResNet101 & 79.48 & 93.47 & 77.44   & 70.93       & 80.91  & 54.13 & 90.16 & \textbf{75.21} & 1201  & 74.80 & 58         \\
    \hline
    \textbf{HoughLane-S}   &   ResNet18       &   78.73    &   92.91    &  77.29     &    73.02 &         78.47  &   52.97    & 89.14  &  60.67 &  1299 &  74.06  &  155             \\
    \textbf{HoughLane-M}   & ResNet34        &  78.87   &   93.04   &  77.43  &   \textbf{73.03}  &  77.69  &  53.03  &  89.36  &  62.88  &  1242  &  74.37  &      148       \\
    \textbf{HoughLane-L}  & ResNet101         &  78.97  &  92.78  &  77.74  &  70.25           &    79.95   &       54.46    &      89.42    &      63.77     &  1350   &  73.79    &       56       \\ 
    \textbf{HoughLane-L}  & PVTv2-b3         &  \textbf{79.81}   &  \textbf{93.63} &  \textbf{78.80}  & 72.66   &  \textbf{82.39}  &   \textbf{55.78}  &  \textbf{90.58}  &  64.92 & 1568  &  \textbf{75.03}    &   61      \\ 
    \hline
    \end{tabular}
    }
    \label{table:test_culane}
    % \vspace{-1em}
\end{table*}

\begin{table}[t!]
    \centering
    \captionof{table}{Comparison of the performance of different methods on LLAMAS. The best results for each metric are given in bold, and F1 is computed using the official CULane source code. Results are tested with FPN, while other necks can also be used.\\}
      \resizebox{\columnwidth}{!}{%
    \begin{tabular}{lccc}
    \hline
    \textbf{Method} & \textbf{Backbone}  & \textbf{Valid-F1$\uparrow$} & \textbf{Test-F1$\uparrow$} \\ \hline
    PolyLaneNet~\cite{tabelini2020polylanenet} & EfficientnetB0 & 90.2 & 88.40 \\
    LaneATT~\cite{LaneATT} & ResNet18 & 94.64 & 93.46 \\
    LaneATT~\cite{LaneATT} & ResNet34 & 94.96 & 93.74 \\
    LaneATT~\cite{LaneATT} & ResNet122 & 95.17 & 93.54 \\
    % CLRNet~\cite{zheng2022clrnet} & ResNet18 & 96.96 & 96.00 \\
    % CLRNet~\cite{zheng2022clrnet} & DLA34 & 97.16 & 96.12 \\
    \hline
    \textbf{HoughLane} &  ResNet18  & 95.62 &   94.75  \\
    % epoch=19, th=0.1, nms, F1=95.62, mF1=62.78
    % epoch=20, th=0.1, nms, F1=95.23, mF1=63.93
    
    \textbf{HoughLane} &  ResNet34  & 95.55 &   94.98  \\
    % epoch=19, th=0.1, nms, F1=95.55, mF1=63.29
    
    \textbf{HoughLane} & ResNet101  & 95.95 &  95.00   \\ 
    % epoch=20, th=0.1, nms, F1=95.82, mF1=64.00
    % epoch=20, th=0.15, nms, F1=94.95, mF1=63.62
    % epoch=18, th=0.1, nms, F1=95.59, mF1=64.59
    % epoch=17, th=0.1, nms, F1=95.95, mF1=65.55
    \hline
    
    \textbf{HoughLane} & PVTv2-b0  & 96.08 &  95.17   \\ 
    % epoch=16, th=0.15, nms, F1=96.08, mF1=61.90
    % epoch=18, th=0.15, nms, F1=95.88, mF1=62.34
    
    \textbf{HoughLane} & PVTv2-b1  & 96.25 &   95.34  \\ 
    % epoch=20, th=0.15, nms, F1=96.25, mF1=63.72
    
    \textbf{HoughLane} & PVTv2-b3  & \textbf{96.31} &  \textbf{95.49}   \\ 
    % epoch=20, th=0.1, nms, F1=96.11, mF1=64.08
    % epoch=20, th=0.15, nms, F1=95.97, mF1=64.06
    % epoch=19, th=0.1, nms, F1=96.30, mF1=65.56
    % epoch=18, th=0.1, nms, F1=96.21, mF1=66.08
    % epoch=16, th=0.1, nms, F1=96.31, mF1=65.83
    \hline
    \end{tabular}
    }
    \label{table:test_llamas}
\end{table}

\subsection{Comparison with Previous Methods}

We compare our method with state-of-the-art methods on the TuSimple, LLAMAS and CULane datasets. Figure~\ref{fig:example_lane} shows some visualization results of detected lanes with different methods on TuSimple and CULane datasets, where each lane is represented by one color. Tables~\ref{table:test_tusimple}, \ref{table:test_culane} and \ref{table:test_llamas} 
quantitatively compare the results of our method with other methods, respectively.

% \noindent
\textbf{Comparison on TuSimple.} 
Quantitative comparison results on the TuSimple dataset are shown in Table~\ref{table:test_tusimple}. Since the TuSimple dataset is small and only contains images of highways, different methods can achieve good F1-scores with small gaps between them. Compared with existing state-of-the-art methods, our method achieves an F1-score of $97.68$ and a best accuracy score of $96.93$. % Moreover, \zjq{compared with the same small backbone ResNet18 of method CondLaneNet~\cite{condlanenet},} our method makes a significant improvement with better gathered local features of lanes and segments.

% \dhb{
% \noindent
\textbf{Comparison on CULane.}
Quantitative comparison results on the CULane dataset are given in Table~\ref{table:test_culane}. Our method achieves an overall F1-score of $78.73$ and an FPS of $155$ with ResNet18, and an F1-score of $79.81$, an FPS of $61$ with the PVTv2-b3 backbone. Although our method does not perform the best on lanes with high curvature due to its line-based feature aggregation prior, it performs better on lane images of diverse scenarios, especially on crowded, dazzle and shadow scenes, where the visual features of lanes are not as strong as those in other scenarios. Further analysis of different backbones is discussed in supplementary material.
% Section~\ref{sec:ablation_backbone}.

We visualize different prediction results of various methods on the TuSimple and CULane datasets in Figure~\ref{fig:example_lane}. We note that our method is able to predict lanes with outlier endpoints, while previous methods only predict lanes with similar endpoints, as shown in the first row. This is because our feature aggregation better captures the subtle features of slim lines and projects lane features into the Hough parameter space, where the selection algorithm determines lane instances based on the distribution of Hough map, without the dependency of point features in the original image space. Our method also better fits the curvature at the top of lanes. For instance, the second row and third row demonstrate that our method successfully predicts the rightmost lane while not extending the lanes too much such that it deviates from the ground plane. In addition, our method consistently predicts correct lanes in dazzle or night scenarios, as shown in the last three rows. This illustrates the effectiveness of our Hough-based feature aggregation based on the line prior.

% \noindent
\textbf{Comparison on LLAMAS.}
Quantitative comparison results on the LLAMAS dataset are given in Table~\ref{table:test_llamas}. Our method achieves an F1@50 of $96.31$ on the validation set, and an F1@50 of $95.49$ on the test set with PVTv2 backbone, which outperforms PolyLaneNet~\cite{tabelini2020polylanenet} and LaneATT~\cite{LaneATT} substantially by 7.09 F1@50 and 1.75 F1@50 on the test set.
% }

\begin{table}[t!]
\begin{center}
\caption{Ablation study of components in HoughLaneNet on CULane.\\}
  \resizebox{\columnwidth}{!}{%
\begin{tabular}{cccccr}
\hline
\textbf{Baseline} & \textbf{S-DHT}  & \textbf{M-DHT} & \textbf{H-Decoder}& \textbf{RHT} & \textbf{F1$\uparrow$} \\ \hline
$\surd$ &   &   &   &   & 70.56    \\
& $\surd$ &   &   &   &   73.11  \\
& $\surd$ &  &  $\surd$ &   & 75.12    \\
% & & $\surd$ &   &   & 73.31    \\
&   & $\surd$ & $\surd$ &   &  77.89  \\
% &   & $\surd$ & & $\surd$ &   \\
&   & $\surd$ & $\surd$ & $\surd$ &  \textbf{78.73}  \\
\hline
\end{tabular}
}
\label{table:components}
\end{center}
\end{table}

\subsection{Ablation Study}\label{sec:ablation}

To evaluate the various aspects of our proposed model in detail we conduct an ablation study. We begin by assessing the effectiveness of major components in our network, then analyze the influence of the image features extracted by different backbones based on ResNet and Pyramid Vision Transformer on the detection results. Finally, we analyze the impact of various key hyperparameters and input image resolutions on the test results. These experiments provide a thorough analysis of the effectiveness of our network framework and demonstrate the rationality of our hyper-parameter selection. Because of space limitations, only a comparison of different components is presented here. For comprehensive details regarding the ablation study, kindly refer to the supplementary material.

To investigate the effectiveness of the proposed components in the network, we report the component-level ablation study in Table~\ref{table:components}. Starting from the baseline that directly regresses the parameters of each instance with a simple convolution module, we progressively include a single-layer DHT module, multi-scale DHT module, RHT line loss, and the upscaled map decoder module. Here, the baseline network retains the dynamic convolution structure besides the backbone and FPN, which is more effective in distinguishing different lanes compared to outputting multiple channels. Our single-layer DHT module (S-DHT) first concatenates the first three layers of features outputted by the neck, then uses Deep Hough Transform to convert it to parameter space. S-DHT improves the baseline results drastically from an F1-score of $70.56$ to $73.11$. This result shows the high effectiveness of our feature aggregation module based on Deep Hough Transform.

In addition, We upscaled the original Hough feature to a 3x map with the map decoder (H-Decoder) to select lane instance points more accurately. This enhancement has further raised the F1-score to $75.12$. The multi-scale DHT module (M-DHT) converts the three layers of features using three separate Deep Hough Transform modules, where each DHT module outputs a scaled Hough feature which matches the input feature size. After that, three multi-scale Hough features are upscaled to the same size and combine with each other with concatenation and a convolution module. With the multi-scale feature gathered by this hierarchical structure, M-DHT improved the F1-score to $77.89$. The addition of RHT loss further improves the F1-score from $77.89$ to $78.73$, which verifies the effectiveness of the inverse Hough transform in improving the supervision accuracy of feature localization.

\section{Conclusion}

In this paper, we introduce a novel lane detection framework utilizing the Deep Hough Transform. This framework aggregates features on both a local and global scale, promoting line-based shapes. It identifies lane instances based on positional differences without relying on pre-defined anchors. A dynamic convolution procedure is employed for pixel-level instance segmentation. To enhance the prediction accuracy, we also incorporate a Hough decoder and additional supervision through a reverse Hough transform loss. This helps to minimize any potential inaccuracies in the construction of the ground-truth Hough points. We evaluate our network on three public datasets and show that our network achieves competitive results, with the highest accuracy of up to $96.93\%$ on TuSimple, a high F1-value of $95.49$ on LLAMAS, and an F1 value of $79.81$ on CULane.

% \noindent 
\textbf{Limitations and Future Work.} The Deep Hough Transform has the capability to extract the features of lane markings by leveraging the line shape geometry prior, but it may not be effective in cases where the lanes have significant curvature or when there is limited line shape information available. Our experiments showed that some of the predicted Hough points were not precisely located near the ground-truth points, resulting in a deviation in the final predicted lane orientation. Despite the implementation of the Reverse Hough Transform output line map and the addition of a constraint loss function to refine the Hough Transform output, the accuracy of predictions remains suboptimal in these scenarios.

Studying the performance impact of various components in our framework reveals three critical factors that support the high accuracy of our network: using the multi-scale features extracted by the backbone and neck; choosing the appropriate size of the transformed Hough map; training with the precise position of predicted Hough points. On the one hand, as the Hough transform-based feature aggregation produces the clearest Hough points when the lane is a perfect line, it usually has degraded performance on lanes with higher curvature. On the other hand, the position of Hough points can be directly regressed to reduce the dependency on a threshold value in the existing point selection strategy. More studies can fruitfully explore these issues further by changing the definition of the parameter space.

%%Vancouver style references.
\bibliographystyle{cag-num-names}
\bibliography{Alane_cag}

\end{document}